\documentclass[letterpaper, 10 pt, conference]{ieeeconf}  

\IEEEoverridecommandlockouts                              

\title{\LARGE \bf
FineGrasp: Towards Robust Grasping for Delicate Objects
}
\usepackage{graphicx}
\usepackage{amsmath}
\usepackage{amssymb}
\usepackage{booktabs}
\usepackage{algorithm}
\usepackage{algorithmic}
\usepackage{multirow}
\usepackage{colortbl}
\usepackage[table]{xcolor}
\usepackage{hyperref}
\hypersetup{
    colorlinks=true,
    linkcolor=cyan!60!blue,
    urlcolor=cyan!60!blue,
    citecolor=green
}

\author{Yun Du$^{*}$, Mengao Zhao$^{*}$, Tianwei Lin, Yiwei Jin, Chaodong Huang, Zhizhong Su
\thanks{* These authors contributed equally.}
\thanks{All authors are with Horizon Robotics.}%
}

\begin{document}

\maketitle
\thispagestyle{empty}
\pagestyle{empty}

\begin{abstract}


Recent advancements in robotic grasping have led to its integration as a core module in many manipulation systems.
For instance, language-driven semantic segmentation enables the grasping of any designated object or object part.
However, existing methods often struggle to generate feasible grasp poses for small objects or delicate components, potentially causing the entire pipeline to fail.
To address this issue, we propose a novel grasping method, FineGrasp, which introduces improvements in three key aspects.
First, we introduce multiple network modifications to enhance the model’s ability to handle delicate regions.
%
Second, we address the issue of label imbalance and propose a refined graspness label normalization strategy.
Third, we introduce a new simulated grasp dataset and show that mixed sim-to-real training further improves grasp performance.
%
%
Experimental results show significant improvements, especially in grasping small objects, and confirm the effectiveness of our system in semantic grasping. Code will be available at \href{https://github.com/HorizonRobotics/robo_orchard_lab/tree/master/projects/finegrasp_graspnet1b}{robo\_orchard\_lab/finegrasp}.

\end{abstract}

\section{INTRODUCTION}

\begin{figure}[t]	
    \begin{minipage}[b]{1\linewidth}
        \centering
        \centerline{\includegraphics[width=8cm]{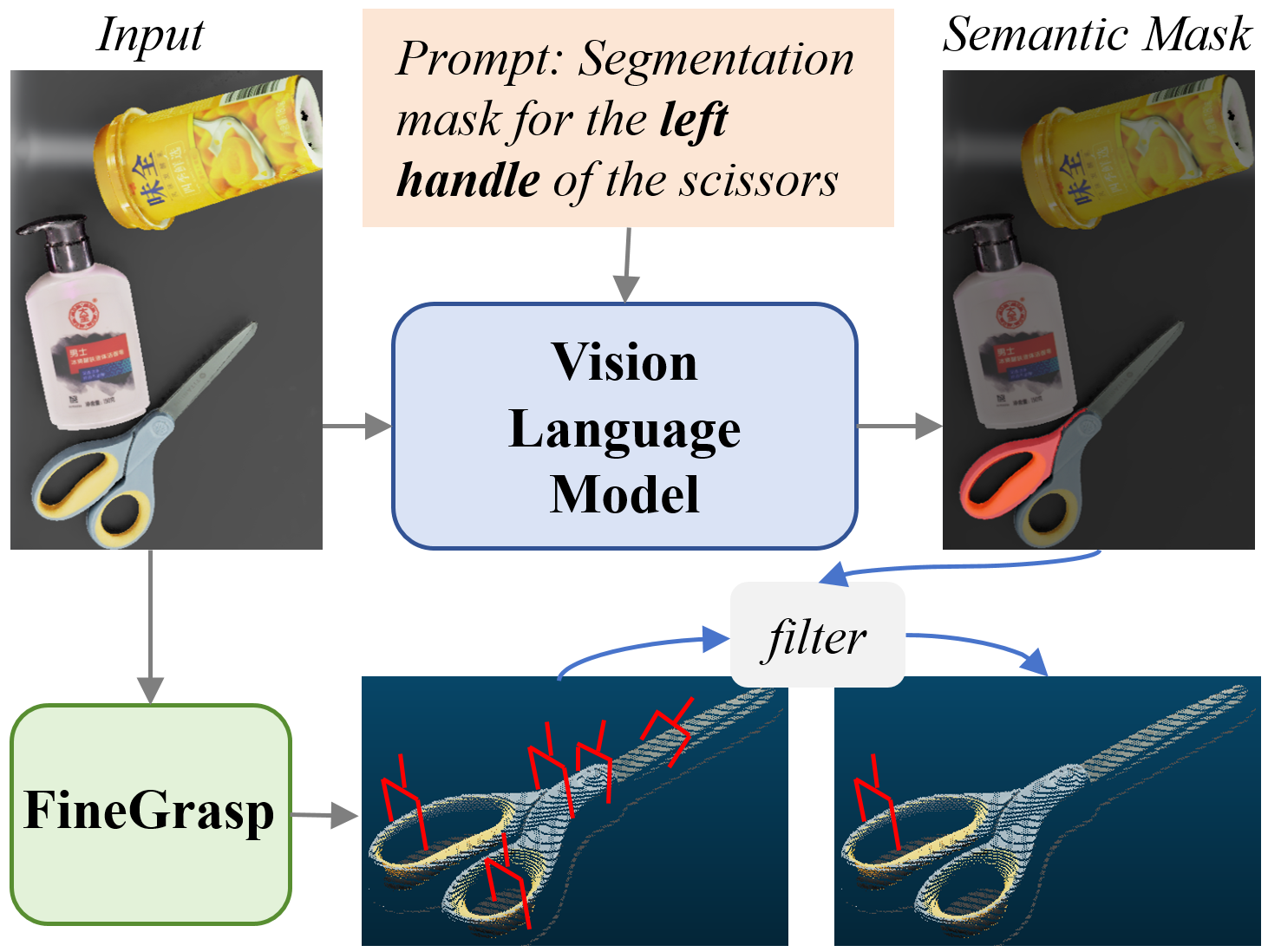}}
    \end{minipage}
    \caption{
Overview of a common semantic grasping framework. First, a pre-trained Vision-Language Model (VLM) is utilized to locate the desired (part of) object. Next, a point-cloud based grasping module  generates the corresponding grasp poses. 
The primary challenge here lies in the difficulty of generating grasp poses for small objects and dedicate components.
}
    \label{fig:overview}
\end{figure}

As one of the most frequently used end-effector designs for robotic arms, the two-finger parallel grippers have driven the development of dedicated 6-DoF grasp detection task.
Recent data-driven methods~\cite{fang2020graspnet,fang2023anygrasp,wu2024economic} have demonstrated remarkable performance, making them a fundamental component of many robotic manipulation systems.
%
%
As shown in Fig. \ref{fig:overview}, a prevalent framework  integrates the grasp detection module with a semantic reasoning module (e.g., GPT-4o, SAM), which generates masks to filter grasp poses for objects specified by language commands. This semantic grasping capability enables a variety of tasks, including task-oriented grasping~\cite{huang2024copa} and part grasping~\cite{qian2024thinkgrasp}.

%
%

Existing grasping methods~\cite{fang2020graspnet,wu2024economic} have demonstrated robust performance for various objects with regular shapes and size, achieving an average grasp success rate exceeding 90\%.
%
However, as illustrated in Fig. 2(a), our implementation of the semantic grasping system reveals that the grasp model occasionally fails to generate feasible grasp poses for small objects or small parts of medium and large objects, leading to the failure of overall system.
%
%
In \cite{Ma_2022_CoRL}, the scale of grasp pose is defined as the minimum width between the gripper’s fingers. Accordingly, delicate grasping issues can be attributed to small-scale grasp poses.


Why do existing methods struggle with small-scale grasp pose estimation? We believe the key issue is that during training, networks tend to focus on larger, easier-to-grasp objects in complex scenes, leading to the neglect of finer objects. Prior work \cite{Ma_2022_CoRL} discussed this  scale imbalance problem, and addressed them by incorporating multi-scale features, scale-aware weighted loss, and explicit balanced sampling via an object segmentation network during inference. 
%
%
However, despite their initial effectiveness, these improvements are now insufficient and come with added computational overhead.

%
%

\begin{figure*}[thb]	
    \begin{minipage}[b]{1\linewidth}
        \centering
        \centerline{\includegraphics[width=18cm]{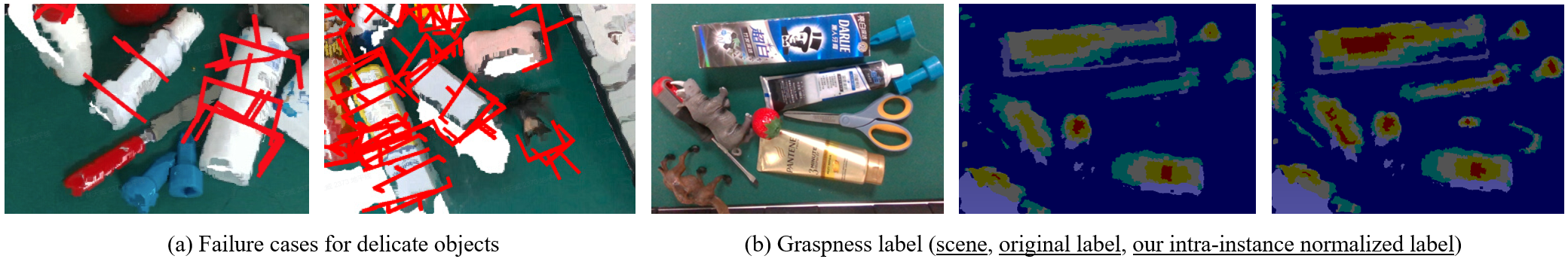}}
    \end{minipage}
    \caption{
Illustrating the challenges of delicate object grasping based on EconomicGrasp:
(a) Two typical failure scenarios: the left image shows the model failing to generate a usable grasp pose for small objects in a cluttered scene, while the right image shows poor pose quality, leading to potential failure.
(b) Issues in graspness ground truth generation: the middle image shows ground truth after cross-object normalization, where scissors are ignored. The right image shows our strategy, normalizing within each object first to ensure consistent scoring across objects.
}
    \label{fig:analysis}
\end{figure*}

In this work, we propose a novel method \textbf{FineGrasp}, to address these challenges with three key improvements.
\textbf{First}, existing methods select sparse points (e.g., 1024) from the dense point cloud based on graspability scores for grasp poses estimation. However, as shown in Fig. 2(b), global normalization of graspability labels often suppresses those of small objects, making them harder to learn.
To overcome this, we introduce a new strategy that normalizes labels within each object first.
This ensures that small objects receive effective supervision during training, and eliminates the need for a segmentation network during inference. 
\textbf{Second}, we enhance existing methods \cite{wu2024economic,Ma_2022_CoRL} by introducing  a multi-range attention module that extracts and fuses multi-scale features around candidate sample points via attention, allowing the model to learn scale-specific weights. We also integrate object surface normals as an auxiliary input to the grasping model, leveraging their strong correlation with optimal grasping directions to improve the network’s accuracy in orientation estimation.
\textbf{Third}, depth sensor noise may lead to  small objects being overlooked during training. To address this, we create supplementary simulation scenes using assets from GraspNet\cite{fang2020graspnet}, which are combined with real data for mixed training. This strategy effectively improves overall performance.
In summary, our contributions are as follows:

\begin{itemize}
\item We propose FineGrasp, a novel method that improves grasp pose accuracy for fine objects, demonstrating significant performance gains in experiment. 
\item We address the fine object grasping problem by introducing a new graspability ground truth generation strategy and several model improvements.
\item We create an extended simulation dataset for GraspNet, using mixed training to enhance performance.
\item We introduce a VLM-based semantic grasping framework to validate the effectiveness of our method.
\end{itemize}

\section{RELATED WORK}

\subsection{6-DoF Grasp Pose Detection}


The 6-DoF grasp pose detection task revolves around two fundamental questions: "Where to grasp?" and "How to grasp?" Accordingly, most existing methods adopt a two-stage prediction pipeline — first, point sampling is used to identify graspable regions, followed by grasp pose prediction in the second stage.
GSNet \cite{wang2021graspness}  introduced a graspness prediction task during the point sampling phase, enabling the model to more effectively identify graspable regions in cluttered scenes. 
To tackle scale imbalance, Scale Balanced Grasp \cite{Ma_2022_CoRL} proposed an object-balanced sampling strategy using an auxiliary segmentation network, which significantly improves performance on small-scale grasps. Additionally, EconomicGrasp \cite{wu2024economic} identified dense supervision as a bottleneck in existing grasp models and proposed an economic supervision paradigm, achieving superior grasp performance with reduced computational cost.


\subsection{Grasp Datasets}
GraspNet-1Billion \cite{fang2020graspnet} is the most extensively utilized dataset for two-finger gripper grasps. It comprises 190 cluttered scenes constructed in real-world environments. A robotic arm equipped with an RGB-D sensor was employed to capture images from various perspectives, while the 6-DoF grasp poses were acquired through analytical computation of force closure. Nevertheless, data collection in real-world settings is inherently time-consuming and challenging, which hinders efforts to achieve large-scale expansion. As a result, certain methodologies advocate for the synthesis of grasp data within a simulation environment. For example, Sim-Grasp \cite{li2024sim} introduced an approach-based sampling scheme coupled with dynamic evaluation in Isaac Sim simulator \cite{liang2018gpu} to generate  grasping labels, and encompasses a substantial number of open-source assets. Furthermore, several methodologies extend this into the domain of dexterous hand grasping, such as DexGraspNet 2.0 \cite{zhang2024dexgraspnet}.

\subsection{Grasp Model in robotic manipulation}
As a fundamental component of real-world robotic manipulation, various approaches have been developed to integrate grasp models into systems, aiming to improve efficiency and precision in object handling.
OK-Robot \cite{liu2024ok} proposed a pick-and-drop solution for real-world applications by integrating VLMs for object detection, utilizing navigation primitives for movement, and employing grasping primitives for object manipulation. In this framework, AnyGrasp \cite{fang2023anygrasp} was used  to generate grasp poses.
CoPa \cite{huang2024copa}  leveraged the common sense knowledge embedded within foundation models to enable the task-relevant object manipulation and utilized GraspNet \cite{fang2020graspnet} as a key grasping module to generate 6-DOF grasp poses.
%
%
To enable manipulation in unstructured environments, OmniManip \cite{pan2025omnimanip} proposed an object-centric primitives framework to better utilize VLMs, where AnyGrasp \cite{fang2023anygrasp} is employed.

\begin{figure*}[thb]	
    \begin{minipage}[b]{1\linewidth}
        \centering
        \centerline{\includegraphics[width=16cm]{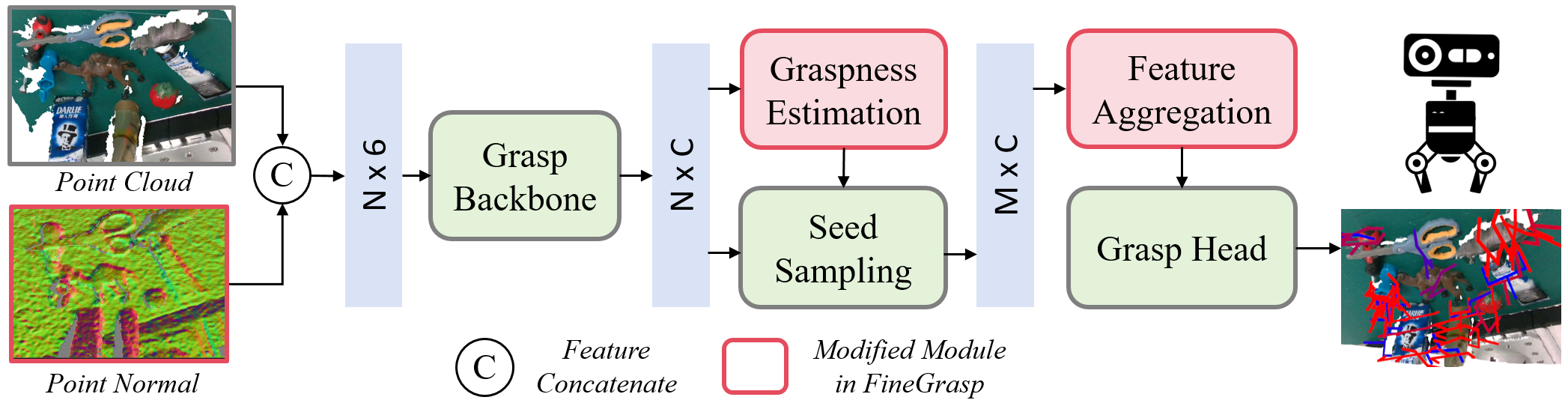}}
    \end{minipage}
    \caption{
Overall framework of our proposed FineGrasp. Based on  EconomicGrasp \cite{wu2024economic}, we  introduce three key improvements: (1) Instance-normalized graspness labels to better balance different objects and ensure delicate objects are not overlooked during seed sampling; (2) Multi-range feature attention module more effective aggregation of multi-scale features. (3) Normal prior as an input to guide the network in identifying the optimal grasping orientation.
}
    \label{fig:Framework}
\end{figure*}

\section{Method}

In this section, we present our method in detail. 
First, in Sec. \ref{method_pre}, we define 6-DoF grasp detection and outline the typical framework of existing grasp models.
Next, in Sec. \ref{sec_method_model}, we introduce our model improvements, including intra-instance graspness label normalization strategy, multi-range attention module, and normal prior.
In Sec. \ref{sec_method_data},  we describe the process of constructing our SimGraspNet dataset.
Finally, in Sec. \ref{sec_method_semantic}, we introduce our semantic grasping solution, which leverages the capabilities of the VLM model for improved applicability in real-world scenarios.



\subsection{ Preliminary}
\label{method_pre}

Consistent with previous studies, we address the challenge of 6-DoF grasp detection for parallel grippers in cluttered scenes, utilizing a single-view RGBD image as input. The grasp pose is represented as \(g = (x,y,z, \theta, \gamma, \beta, w )\), where \((x, y, z)\)  denotes the center of grasp, w is the grasp width and  ($\theta$, $\gamma$, $\beta$) is the intrinsic rotation as Euler angles.

Typical grasp models~\cite{fang2020graspnet, wu2024economic} are generally comprised of three integral modules. 
\textbf{(1)} Initially, a 3D convolutional backbone extracts geometric features from the input point cloud, while an MLP block estimates each point's graspability. We then select $K$ points based on this graspability.
\textbf{(2)} Subsequently, for each sampled point, an MLP block selects the optimal view, and the cylinder grouping module arranges local geometric features accordingly within a cylindrical space.
\textbf{(3)} Finally,  an MLP block forecasts the grasp scores, grasp depths, as well as the grasp widths for each grouping. 

EconomicGrasp \cite{wu2024economic} further  introduces an economic supervision paradigm, optimizing label selection and training efficiency to overcome the dense supervision bottleneck. Building upon this strong baseline, we have undertaken efforts to improve the grasp performance of  delicate objects.


\subsection{Grasp Model}
\label{sec_method_model}

\noindent
\textbf{Instance-Norm Graspness.}
%
As previously discussed, identifying graspable regions is crucial for the success of grasping methods. While \cite{wang2021graspness} suggests that scene-level graspness prediction enhances the selection of suitable seed points, we identify two key limitations: (1) Grasp poses on delicate objects are often penalized for collisions, leading to artificially low graspness scores; (2) Scene-level normalization applies a uniform scaling across all objects, disregarding inter-object variations. This global suppression disproportionately lowers the scores of small objects, further hindering their learning.

\begin{equation}
\tilde{S}_P^{\prime}  = \left\{ \frac{\tilde{s}^c_p - \min(\tilde{\mathcal{S}}^c_p)}{\max(\tilde{\mathcal{S}}^c_p) - \min(\tilde{\mathcal{S}}^c_p)} \ \bigg| \ c \in \mathcal{C} \right\}
\label{eq:eq1}
\end{equation}


 \begin{equation}
S_P = \left\{ \frac{\tilde{s}^{\prime~i}_p - \min(\tilde{\mathcal{S}}_p^{\prime})}{\max(\tilde{\mathcal{S}}_p^{\prime}) - \min(\tilde{\mathcal{S}}_p^{\prime})} \ \bigg| \ i = 1, \ldots, N \right\}
\label{eq:eq2}
\end{equation}

To address these issues, we introduce Instance-Normal Graspness, which aims to balance the learning of graspable regions across objects of different sizes.
Following the Graspness generation process of GSNet \cite{wang2021graspness}, we obtain  graspness scores for each of $N$ points in the scene, indicating their graspability. 
GSNet uses global scene-level normalization for graspable supervision. 
In contrast, we first normalize graspness scores within each category c to highlight graspable areas within objects. Here, $\mathcal{C}$ is the set of object categories, and $\tilde{s}^c_p$ denotes the graspness score (with dimension $N_c\times 1$) for points in category $c$ , where $N_c$ is the number of points in the category. 
This instance-level normalization is shown in Eq.\ref{eq:eq1}. Finally, we apply the scene-level normalization to obtain the final scene-level graspness $S_P$ as shown in Eq.\ref{eq:eq2}.
  



\begin{figure}[t]	
    \begin{minipage}[b]{1\linewidth}
        \centering        \centerline{\includegraphics[width=9cm]{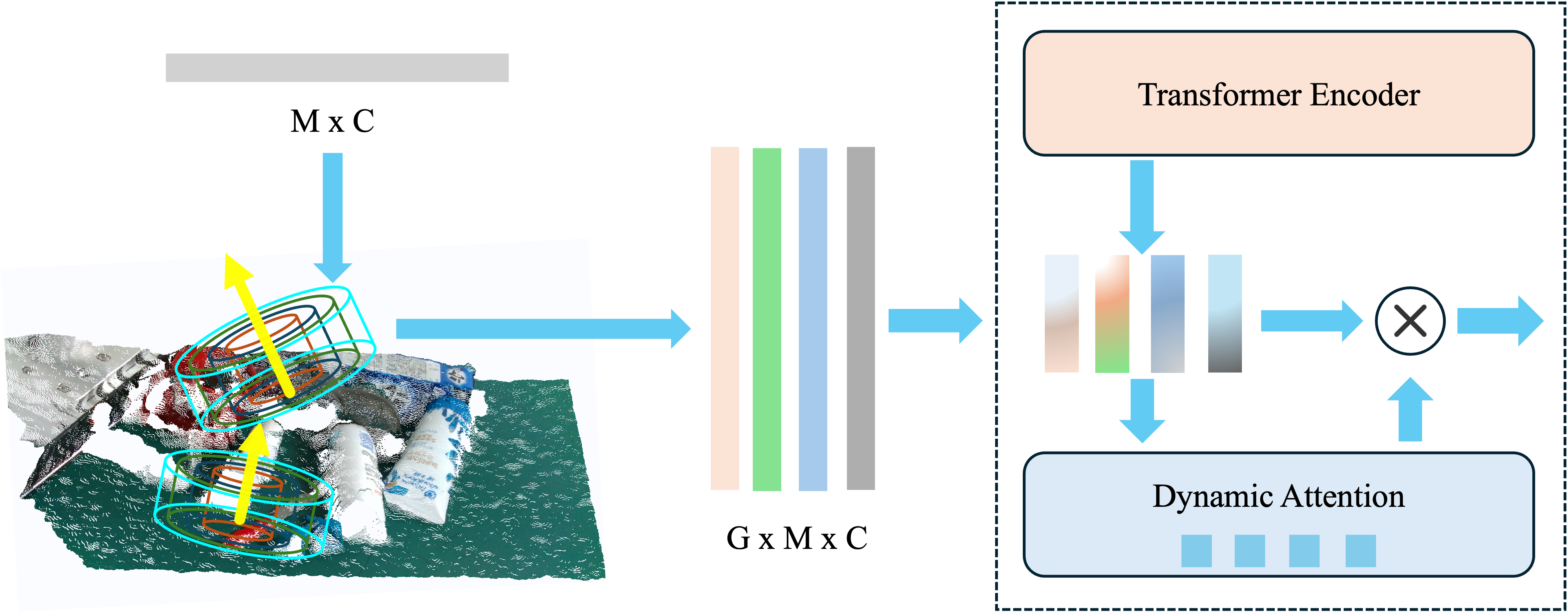}}
    \end{minipage}
    \caption{Point features are aggregated across multiple ranges through a Transformer encoder, enabling cross-scale feature interaction. Adaptive fusion weights dynamically combine these features, facilitating grasp pose learning for varying sizes.}
\end{figure}

\noindent
\textbf{Multi-range Attention.}
The graspability of an object is highly correlated with the features within its local surroundings. Therefore, for feature aggregation of each candidate seed point, it is common practice to aggregate features within a defined receptive field using cylinder grouping.
Scale-Balance-Grasp~\cite{Ma_2022_CoRL} explored multi-scale feature aggregation by employing multiple receptive fields.
Building on this, we introduce a Multi-Range Attention (MRA) mechanism that enables dynamic cross-scale feature interaction, further enhancing feature representation for grasping tasks.

As shown in Fig.\ref{fig:Framework}, after passing through the grasp backbone and seed sampling module, the point cloud yields features with a dimension of $M\times C$, where $M$  and $C$ represent the sample point number and feature dimension respectively. We set $G$ radius ranges and apply cylinder grouping for local feature aggregation. The cylinders are oriented along the view direction, producing a multi-range feature $X \in \mathbb{R}^{G\times M\times C}$.

\begin{equation}
    F = \text{TransformerEncoder}(X) 
\label{eq:transformer}
\end{equation}

\begin{equation}
O = \sum_{g=1}^G F_g \odot \text{softmax}(W F_g)  
\label{eq:fusion}
\end{equation}

This feature $X$ undergo dimensional transformation and contextual encoding through a Transformer module (Eq.\ref{eq:transformer}), producing latent representations that capture inter-group dependencies. 
Moreover, our network learns adaptive fusion weights via learnable projections (Eq.~\ref{eq:fusion}), ensuring scale-aware feature fusion for pose regression.

\begin{figure}[t]	
    \begin{minipage}[b]{1\linewidth}
        \centering
        \centerline{\includegraphics[width=8cm]{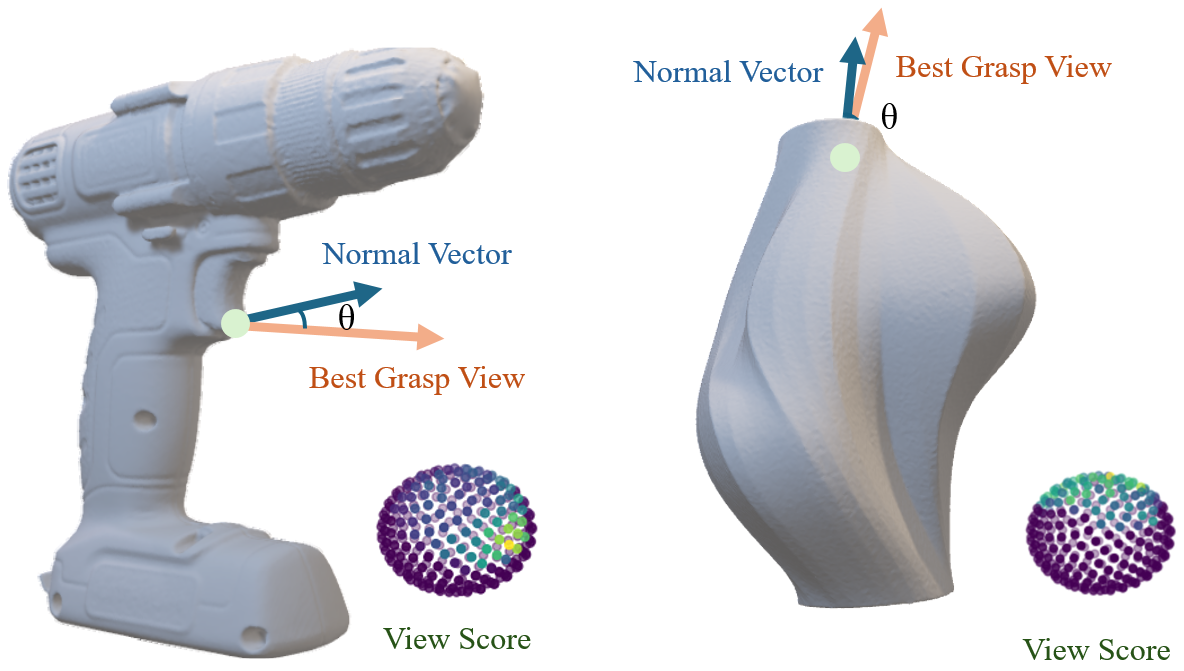}}
    \end{minipage}
    \caption{The force closure score distribution in GraspNet1B exhibits a view-dependent bias, the normal vector provides a approaching prior for high-quality grasping poses regression}
    \label{fig:NormalPrior}	
\end{figure}

\noindent
\textbf{Normal as Approaching Prior.}
%
%
The orientation of a grasp pose can be defined by a combination of a view vector and an in-plane rotation angle. As demonstrated in EconomicGrasp~\cite{wu2024economic}, viewpoint selection significantly impacts grasp pose quality. By averaging grasp scores across various predefined views, we observe in Fig.~\ref{fig:NormalPrior} that certain viewpoints inherently lead to more accurate grasp pose regression.

Building on this observation, we propose that when the grasp pose is aligned with the normal direction of the object, a more stable grasp can be achieved. As shown in  Fig.\ref{fig:NormalPrior}, viewpoints within a $15^{\circ}$ cone of the surface normal direction can on average generate high-quality grasp poses (defined as the top 1\% of all grasp poses). 
To leverage this insight, we compute the normal vector $N\in \mathbb{R}^{N\times 3}$ for each point $p_i$ in the point cloud $P\in \mathbb{R}^{N\times 3}$ based on its neighboring points $p^{right}$, $p^{left}$, $p^{down}$, and $p^{up}$. The normal vector $n_i$ is obtained by taking the cross product of the horizontal tangent vector $v_h = p_{right} - p_{left}$ and the vertical tangent vector $v_v = p_{down} - p_{up}$, and then normalizing the result:





\begin{equation}
n_i = \frac{(p^{right} - p^{left}) \times (p^{down} - p^{up})}{|(p^{right} - p^{left}) \times (p^{down} - p^{up})|}.
\label{eq:normal}
\end{equation}
 
Consequently, we incorporate surface normals as point features ($f_i = [x_i, n_i]$, combining position $x_i$ and normal $n_i$) enhances geometric representation modeling capability, improving grasp pose prediction performance.

\subsection{SimGraspNet dataset}
\label{sec_method_data}

\noindent
\textbf{Simulation Setup.}
We construct our simulation dataset using Isaac Sim \cite{liang2018gpu}. A virtual RGB-D camera is mounted on the robotic arm’s wrist, following a predefined motion trajectory. Referring to GraspNet-1Billion \cite{fang2020graspnet}, the collection trajectory encompasses  256 distinct viewpoints distributed over a quarter-sphere. As the robotic arm moves along this trajectory, we systematically capture RGB-D images, object segmentation masks, camera poses relative to a fixed coordinate system, camera intrinsics, and other metadata.

\noindent
\textbf{Assets and Cluttered Scene.}
We utilize the same set of 40 objects from the GraspNet-1Billion \cite{fang2020graspnet} training set and enhance them with real-world physical properties such as gravity, velocity, acceleration, and mass, enabling realistic interactions.
Subsequently, a subset of these objects is then randomly selected and dropped from varying heights with randomized initial positions and orientations. As they descend, the objects collide with each other and the table, resulting in diverse and naturalistic object poses.  
This process generates approximately 400 cluttered scenes, forming our SimGraspNet dataset. Finally, we generate ground truth labels for these scenes following the methodology proposed in  GraspNet-1Billion\cite{fang2020graspnet}.


\subsection{Semantic Grasp}
\label{sec_method_semantic}

In real-world applications, grasping a user-specified object or a specific part of an object is a common requirement, such as the handle of a hammer.
To enable this semantic grasping capability, we integrate the powerful VLM model with FineGrasp.
As shown in Fig. \ref{fig:overview}, our framework consists of three steps: (1) a pre-trained VLM is used to identify the mask of the target object or its specific part, (2) FineGrasp generates grasp poses for the entire scene, and (3) grasp candidates are filtered, retaining only those within the semantic mask.
%

However, directly applying VLM models to part grounding tasks often results in suboptimal performancedue to their limited fine-grained understanding of objects.
To address this, we propose a coarse-to-fine processing pipeline. Given an input RGB image and instructions, we first use the VLM model to ground the target object at the object level. The identified object is then cropped from the image, allowing for fine-grained part grounding to determine the specific graspable region.
This approach enables effective part grounding even in cluttered scenes. We choose Sa2VA-4B \cite{yuan2025sa2va} for its superior performance in grounding tasks and high efficiency in terms of frames per second (FPS).

\section{Experiments}
In this section, we conduct a comprehensive experimental evaluation of FineGrasp. We begin by detailing the experimental setup, followed by a performance comparison against state-of-the-art methods. Next, we perform ablation studies to analyze the impact of each key component. Finally, extensive real-world robot experiments are conducted to validate FineGrasp's effectiveness in grasping delicate objects and enabling semantic grasping.

\subsection{Experiment Setup}

\noindent
\textbf{Dataset and Metrics.}
We evaluate the performance of our method using the widely adopted benchmark GraspNet-1Billion, which comprises 190 cluttered scenes captured in the real-world. We utilize the original division of the dataset into training and testing subsets, consisting of 100 scenes for training. The test sets are further categorized into Seen, Similar and Novel based on the characteristics of the objects present. The RealSense split is utilized for better depth quality. We follow the official evaluation protocol, wherein detected grasp poses are initially filtered through non-maximum suppression, followed by the evaluation of the top 50 grasp poses using the average precision (AP) metric.

\noindent
\textbf{Baseline comparisons.}
%
Our method is built upon the foundation of EconomicGrasp~\cite{wu2024economic}, with several improvements. We compare it with GraspNet-Baseline~\cite{fang2020graspnet}, Scale Balance Grasp~\cite{Ma_2022_CoRL}, HGGD~\cite{chen2023efficient}, GSNet~\cite{wang2021graspness}, AnyGrasp~\cite{fang2023anygrasp}, and EconomicGrasp~\cite{wu2024economic}.

\noindent
\textbf{Implementation Details.}
%
Our model is implemented using the PyTorch framework and trained on 8 Nvidia 4090 GPUs. The Adam optimizer is used, with a batch size of 4 per GPU and an initial learning rate of 0.001. The learning rate follows a cosine decay schedule with a linear warm-up strategy. The training process spans 10 epochs and takes approximately 2 hours to reach convergence.

\subsection{Comparison with the State-of-the-art}
Table \ref{table:ExperimentResult} presents the comparative results of  FineGrasp with other representative approaches on GraspNet-1B Dataset. Despite our focus on improving performance for delicate objects, FineGrasp outperforms other methods across Seen, Similar, and Novel settings. FineGrasp achieves an average AP of 53.97 on the RealSense split. Additionally, by incorporating collision detection, the average AP increases to 55.47, as shown in the final row of Table \ref{table:ExperimentResult}.


\begin{table}[!hbt]
    \caption{Experiment on GraspNet-1B dataset. Showing APs on Realsense split. CD means Collision Detection}
    \scalebox{0.95}{
        \begin{tabular}{c|c|c|c|c}
            \hline
             Method & Seen & Similar & Novel & Average \\
            \hline
               GraspNet-Baseline~\cite{fang2020graspnet} & 27.56 & 26.11 & 10.55 & 21.41 \\
               Scale Balanced Grasp~\cite{Ma_2022_CoRL} & 58.95 & 52.97 & 22.63 & 44.85 \\
               HGGD~\cite{chen2023efficient} & 64.45 & 53.59 & 24.59 & 47.54 \\ 
               GSNet~\cite{wang2021graspness} & 67.12 & 54.80 & 24.31 & 48.74 \\  
               AnyGrasp~\cite{fang2023anygrasp} + CD & 66.12 & 56.09 &	24.81 &	49.01 \\
               EconomicGrasp~\cite{wu2024economic} & 68.21 & 61.19 & 25.48 & 51.63 \\
            \hline
               FineGrasp & 71.67 & 62.83 & 27.40  & 53.97 \\
               FineGrasp + CD & \textbf{73.71} & \textbf{64.56} & \textbf{28.14} & \textbf{55.47} \\
            \hline
        \end{tabular}}
    \label{table:ExperimentResult} 
    \centering
\end{table}

\subsection{Ablation Studies}

\noindent
\textbf{Grasping model.}
In this section, we assess the effectiveness of the proposed modules. First, to evaluate the effectiveness of our MRA module,we replace the feature fusion method with the one proposed in  Scaled Balanced Grasp \cite{Ma_2022_CoRL}. Additionally, we introduce a scaled metric designed to more accurately assess the scale-aware grasping quality within a scene. Specifically, the maximum width of the gripper is partitioned into three intervals: 0-4 cm for Small, 4-7 cm for Medium, and 7-10 cm for Large.  Table   \ref{tab:MRA} present the compare result, demonstrating that the proposed MRA module achieves superior performance across small, medium, and large objects.

\begin{table}[!hbt]
    \caption{Ablation study of multi-range feature fusion.}
    \scalebox{1}{
        \begin{tabular}{c|ccc|c}
            \hline
            Feature fusion & Small & Medium & Large & Average  \\
             \hline
             MsCG~\cite{Ma_2022_CoRL} & 15.87 & 50.52 &  55.51 & 40.63 \\ 
            MRA & \textbf{16.23} & \textbf{50.67} & \textbf{56.92} & \textbf{41.27} \\
            \hline
        \end{tabular}}
    \centering
    \label{tab:MRA} 
\end{table}

Next, we conduct an ablation study on our proposed modules and simulation dataset, with the results summarized in Table  \ref{table:ablation_study} . The results indicate that the Instance-Norm Graspness module significantly improves small-scale performance, increasing it from 11.47 to 14.74. This improvement is attributed to its mitigation of the over-suppression effect caused by global normalization, allowing small-scale objects to receive more adequate learning during the point proposal stage. Furthermore,  incorporating the Normal Prior module enriches  point cloud features by leveraging surface orientation, thereby strengthening geometric modeling capability. This results in consistent performance gains on the Novel split in our experiments. Moreover, the mix training with our SimGraspNet dataset leads to a significant improvement in small-object performance, increasing AP from 16.23 to 18.87. This highlights the impact of depth sensor noise on small objects and demonstrates that incorporating simulation data can effectively mitigate this issue.


\begin{table*}[!hbt]
    \caption{The impact of components in the proposed method: NP (Normal Prior), MRA (Multi-range Attention), and ING (Instance-norm Graspness).}
	\scalebox{1.2}{
		\begin{tabular}{cccc|cccc|ccc}
		      \hline 
		    	 ING & MRA  &NP& SimGraspNet& Seen & Similar & Novel & Average & Small & Medium & Large \\ \hline

                \hline
		    	     &     &&        &   68.13   & 59.94  & 25.32  & 51.13 & 11.47 & 48.02 & 53.76  \\ 
                    $\checkmark$ &   &&   & 69.51 & 60.94 & 25.44 & 51.96 & 14.74 & 48.53 & 54.85 \\                  
                     & $\checkmark$  && & 70.32 & 61.40 & 25.45 & 52.39 & 14.33 & 49.94 & 54.82 \\                  
                      &   &$\checkmark$& & 69.71 & 60.80 & 25.92 & 52.14 & 13.01 & 48.88 & 55.11 \\   \hline      
                    $\checkmark$  & $\checkmark$  && & 70.89 & 61.61 & 25.76 & 52.75 & 14.63 & 49.38 & 55.52 \\ 
                      & $\checkmark$  &$\checkmark$& & 70.71 & 61.27 & 26.58 & 52.85 & 15.40 & 50.02 & 55.23 \\ 
                    $\checkmark$  & $\checkmark$  &$\checkmark$& & \textbf{71.67} & 62.83& \textbf{27.40} & \textbf{53.97} & 16.23 & \textbf{50.67}& \textbf{56.92} \\  \hline
              
 $\checkmark$&  $\checkmark$&$\checkmark$& $\checkmark$& 70.80& \textbf{63.84}& 26.51& 53.71& \textbf{18.87}& 50.61&56.53\\ 
   \hline
	\end{tabular}}
	\centering
	\label{table:ablation_study}
	     \vspace{0cm}
\end{table*}

\noindent
\textbf{Simulation dataset.}
Table \ref{tab:SimulationDataset} displays the results on our SimGraspNet dataset,  wherein the EconomicGrasp is selected as the grasp model, with the sole variable being the training data. The results demonstrate that, when trained exclusively with our SimGraspNet, a satisfactory performance is attainable, namely 59.35 for Seen, 56.74 for Similar and 23.68 for Novel. To mitigate the discrepancy between simulation and reality, we introduce a Gaussian Shift to the simulated depth images to replicate real sensor noise. Furthermore, we employ an off-the-shelf depth restoration method \cite{yang2025depth} to preprocess the test depth images. The corresponding result, denoted as SimGraspNet+Sim2real, indicates a substantial reduction in the sim-to-real gap across all test sets. Moreover, when trained jointly on both simulated and real-world data, denoted as SimGraspNet+GraspNet-1B,  the performance improves by 3.09, 4.29 and 0.8 respectively when compared to the baseline, thereby substantiating the efficacy of our dataset.

\begin{table}[!hbt]
    \caption{The experiment on simulation dataset}
    \scalebox{1.2}{
        \begin{tabular}{c|c|c|c}
            \hline
             Train set& Seen & Similar & Novel  \\
             \hline
               GraspNet-1B& 68.13& 59.94& 25.32\\
 SimGraspNet& 59.35& 56.74& 23.68 \\
 SimGraspNet + Sim2real& 62.47& 58.41&24.16\\
 SimGraspNet+GraspNet-1B& 71.22& 64.24& 26.12\\
        \hline
        \end{tabular}}
    \centering
    \label{tab:SimulationDataset} 
\end{table}

\begin{figure*}[thb]	
    \begin{minipage}[b]{1\linewidth}
        \centering
        \centerline{\includegraphics[width=16cm]{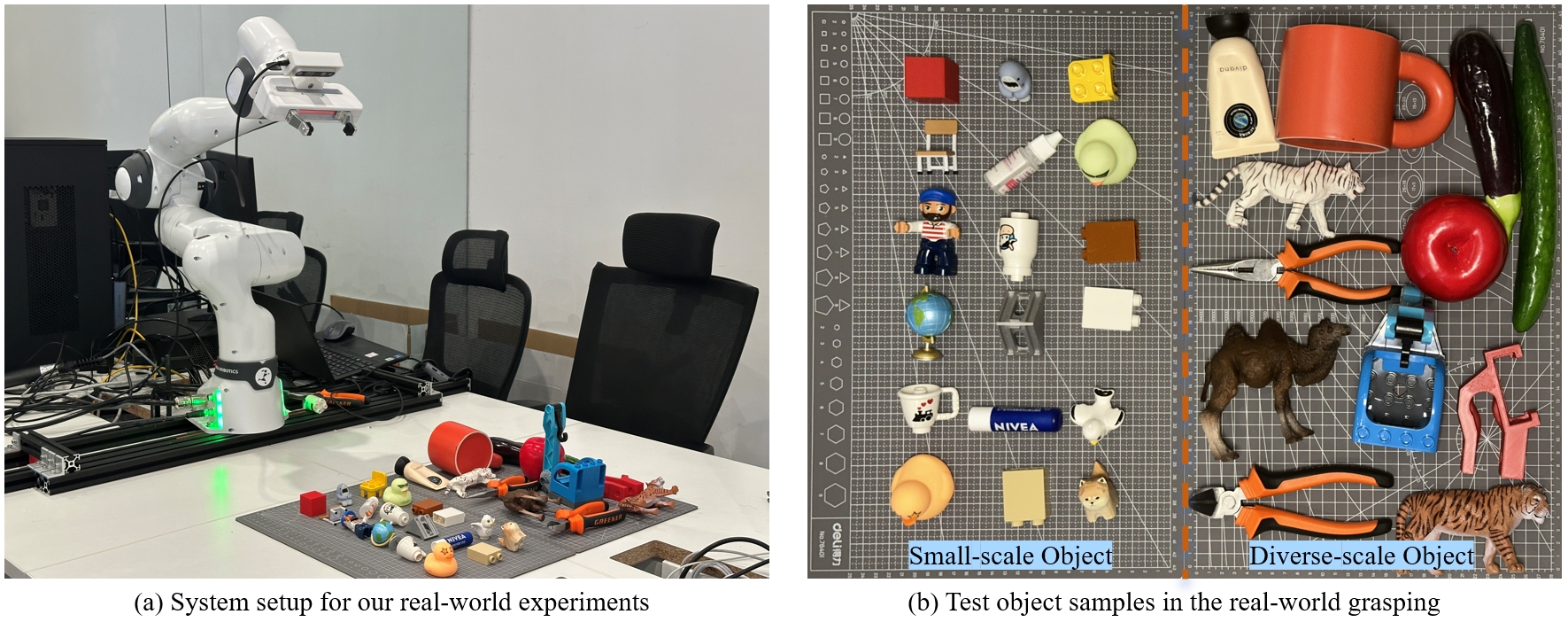}}
    \end{minipage}
    \caption{Robot and object settings in the real-world grasping experiments.}
    \label{fig:setup}
\end{figure*}

\subsection{Real-World Experiment on delicate objects}
We conduct  a real-world experiments to validate the ability of our method, where a 7-DoF FR3 robotic arm, supplied by Franka Emika and equipped with a parallel-jaw gripper, is selected as our platform. Additionally, a RealSense D435i depth sensor is mounted on a tripod positioned in front of the arm as shown in Fig.\ref{fig:setup} (a). A total of 18 small objects were gathered to construct 5 test scenes, with each scene comprising 6 objects, as depicted in Fig.\ref{fig:setup} (b). In each trial, the robotic arm performs the grasping action associated with the highest score, and iteratively removes objects until the workspace is cleared or two failures occur in succession. If two consecutive grasp failures occur, the scene is marked as incomplete.

Table \ref{tab:DelicateObjects} reports the result of real robot experiments. FineGrasp achieves success and completion rates of 91\% and 100\%, respectively, whereas GSNet and EconomicGrasp attain only 78\%/40\% and 69\%/20\%. The substantial gap in completion rate demonstrated by our model underscores its effectiveness in handling delicate object grasps.

\begin{table}[!hbt]
    \caption{Real-world experiment on delicate objects}
    \scalebox{1}{
        \begin{tabular}{cc|ccc}
            \hline
             Scene & Objects & GSNet~\cite{wang2021graspness} & EconomicGrasp~\cite{wu2024economic} & FineGrasp \\\hline
             1 & 6 & 5/6 & 6/6 & 6/6 \\
             2 & 6 & 5/6 & 5/6 & 6/7\\
             3 & 6 & 6/8 & 4/6 & 6/6 \\
             4 & 6 & 2/6 & 4/6 & 6/6 \\
             5 & 6 & 4/6 & 6/8 & 6/8 \\
             \hline
            \multicolumn{2}{c|}{Success Rate} & 69\%(22/32) & 78\%(25/32) & \textbf{91\%(30/33)}\\ 
            \multicolumn{2}{c|}{Completion Rate} & 20\%(1/5) & 40\%(2/5) & \textbf{100\%(5/5)}\\ 
             \hline
        \end{tabular}}
    \centering
    \label{tab:DelicateObjects} 
\end{table}

\subsection{Real-World Experiments on semantic grasp}
To validate our coarse-to-fine semantic grasp system, we conduct empirical grasp experiments under real-world conditions. We utilize two test settings to assess the practical semantic grasping capabilities: Object-grounding grasp and Part-grounding grasp. 
\begin{table}[!hbt]
    \caption{Real-world experiment on object-grounding grasp}
    \scalebox{0.90}{
        \begin{tabular}{ccc|cc}
            \hline
              Scene & Objects & Object Prompt & EconomicGrasp~\cite{wu2024economic} & FineGrasp \\
             \hline
                1 & 8 & Orange duck & 3/3 & 3/3 \\
                2 & 9 & Green box & 3/3 & 3/3 \\
                3 & 12 & globe & 2/3 & 2/3 \\
                 4 & 11 & Lip balm & 0/3 & 2/3\\
                 5 & 10 & Metal workpiece & 3/3 & 3/3\\
             \hline
              \multicolumn{3}{c|}{Success Rate} & 73\%(11/15) & \textbf{86\%(13/15)} \\
             \hline
        \end{tabular}}
    \centering
    \label{tab:ObjectgroundingGrasp} 
\end{table}

In the Object-grounding experiment,  selected delicate objects are placed in a cluster scene containing 8 to 12 objects, with each object positioned randomly. The grasping is repeated three times for each scene. If semantic grounding failure occurs, the trial is excluded from the statistical results.
As shown in the table \ref{tab:ObjectgroundingGrasp}, our method outperforms EconomicGrasp, achieving a success rate of 86\%. Specially, small objects, such as lip balm, in the cluster scene are often overlooked in the point proposal stage, leading to few or no high-quality grasp poses. However, our FineGrasp method achieves a success rate of 66\% for this object, whereas EconomicGrasp fails to complete a single successful grasp.
\begin{table}[!hbt]
    \caption{Real-world experiment on part-grounding grasp}
    \scalebox{0.95}{
        \begin{tabular}{cc|cc}
            \hline
              Object & Part Prompt & GSNet~\cite{wang2021graspness} & FineGrasp \\
             \hline
               Hand  Cream & Cap of the hand cream & 0/5 & 4/5 \\
                Cup & Handle of the cup & 5/5 & 3/5 \\
                Crane &Arm of the crane & 4/5 & 5/5 \\
                 Tiger & Head of the tiger & 4/5 & 5/5\\
                 Pliers & Left handle of the pliers & 1/5 & 4/5\\
             \hline
              \multicolumn{2}{c|}{Success Rate} & 56\%(14/25) & \textbf{84\%(21/25)} \\
             \hline
        \end{tabular}}
    \centering
    \label{tab:PartGroundingGrasp} 
\end{table}

In the Part-grounding grasp setting, we collect 5 objects, each featuring distinct parts, such as cup and pliers, and place them randomly with various orientations. Given a specific instruction, such as "the handle of the cup",  5 attempts were made accordingly. Note that a trial was deemed successful only if the specified part of the object was grasped. As shown in Table \ref{tab:PartGroundingGrasp}, our approach attained an average success rate of 84\%, demonstrating the potential of integrating FineGrasp into downstream manipulation applications.

\section{CONCLUSIONS}


In this work, we propose FineGrasp, a novel method that enhances grasp pose estimation for fine objects through improved label normalization, multi-range attention, normal prior, and mixed training with simulated data. These innovations address key challenges in delicate object grasping, leading to more accurate and robust predictions. Beyond improving grasp performance, our approach eliminates the need for segmentation during inference and offers a more data-efficient training strategy. This work contributes to advancing robotic grasping in cluttered and unstructured environments. 
%
%
Future work could expand simulation scenarios for small and articulated objects to further enhance generalization and robustness of grasp model.

\addtolength{\textheight}{-12cm}   

\bibliographystyle{IEEEtran}
\bibliography{IEEEexample}

\end{document}